\documentclass[11pt]{article}
\usepackage[nonatbib,final]{neurips_2022}
\usepackage[numbers]{natbib}
\pdfoutput=1
\usepackage{graphicx}
\graphicspath{ {images/} }
\usepackage{subcaption}
\usepackage[utf8]{inputenc} 
\usepackage[T1]{fontenc}    
\usepackage{hyperref}       
\usepackage{url}            
\usepackage{booktabs}       
\usepackage{amsfonts}       
\usepackage{nicefrac}       
\usepackage{microtype}      
\usepackage{xcolor}         
\author{han4n}
\date{\today}
\title{}
\hypersetup{
 pdfauthor={han4n},
 pdftitle={},
 pdfkeywords={},
 pdfsubject={},
 pdfcreator={Emacs 28.2 (Org mode 9.5.5)},
 pdflang={English}}
\begin{document}

\title{I see you: A Vehicle-Pedestrian Interaction Dataset from Traffic
  Surveillance Cameras}
%

\author{%
Hanan Quispe$^{1}$\thanks{These authors contributed equally to this work} \quad Jorshinno Sumire$^{1}$\footnotemark[1] \quad Patricia Condori$^2$ \quad Edwin Alvarez$^3$ \\ \quad \textbf{Harley Vera$^1$}\\
$^1$Universidad Nacional de San Antonio Abad del Cusco \quad $^2$Business on Engineering \\ and Technology - BE Tech  \quad $^3$Pontificia Universidad Católica del Perú\\
\texttt{\{163819,160347,harley.vera\}@unsaac.edu.pe}\\
\texttt{patricia.condoriobregon@betech.net.pe}\\
\texttt{edwin.alvarez@pucp.edu.pe}
}

\maketitle

\begin{abstract}
The development of autonomous vehicles arises new challenges in urban traffic scenarios where vehicle-pedestrian interactions are frequent e.g. vehicle yields to pedestrians, pedestrian slows down due approaching to the vehicle. Over the last years, several datasets have been developed to model these interactions. However, available datasets do not cover near-accident scenarios that our dataset covers. We introduce \emph{I see you}, a new vehicle-pedestrian interaction dataset that tackles the lack of trajectory data in near-accident scenarios using YOLOv5 and camera calibration methods. \emph{I see you} consist of 170 near-accident occurrences in seven intersections in Cusco-Peru. This new dataset and pipeline code are available on GitHub\footnote{https://github.com/hvzzzz/Vehicle\_Trajectory\_Dataset}.
\end{abstract}

\section{Introduction}
\label{sec:orgb75ea6f}

The problem of predicting pedestrian paths is crucial for the development of autonomous vehicles(AVs) that have to handle complex interactions with pedestrians in urban traffic environments e.g. pedestrians recklessly crossing the road, and vehicles speeding up to overtake a pedestrian. To address this problem multiple datasets have been developed \cite{yang_top-view_2019,bhattacharyya_euro-pvi_2021,chandra_meteor_2021,bock_ind_2019}. Even though existing datasets cover diverse vehicle-pedestrian interaction scenarios, vehicle-pedestrian interaction in near accident scenarios is especially important to avoid situations that can lead to safety hazards and is challenging because pedestrians exhibit unpredictable maneuvers with sudden speed changes \cite{alhajyaseen_studying_2017}. Those sudden behavioral changes are hard to predict by an AV as verified by \cite{wang_adaptability_2020} where the authors argued that safely maneuvering through jaywalkers is a complex phenomenon that needs attention.

Near accident, data is difficult to collect \cite{wu_novel_2018}. This has led to the development of multiple data-collecting approaches. \citet{wu_improved_2020} examined the usage of a roadside LIDAR sensor to collect trajectories of road users and identify vehicle-pedestrian near-crash events. \citet{lee_high_2016} conducted a study to collect physiological data from night-shift workers in near-crash scenarios where 43.8\% of the drives were terminated early for safety reasons. \citet{nasernejad_modeling_2021} developed an agent-based framework to model pedestrian behavior in near misses and used surveillance camera footage to validate their method. \citet{anik_investigation_2021} collected jaywalkers' trajectories from surveillance camera footage and proposed an artificial neural network to predict jaywalkers' trajectory on a mid-block location in Bangladesh-India. To train or calibrate such models is important to have publicly available ground truth trajectory data. Nevertheless to the best of the author's knowledge, this is the first publicly available trajectory dataset that covers near-accident scenarios in Latin America.

\begin{figure}[htbp]
\centering
\includegraphics[angle=0,width=12cm]{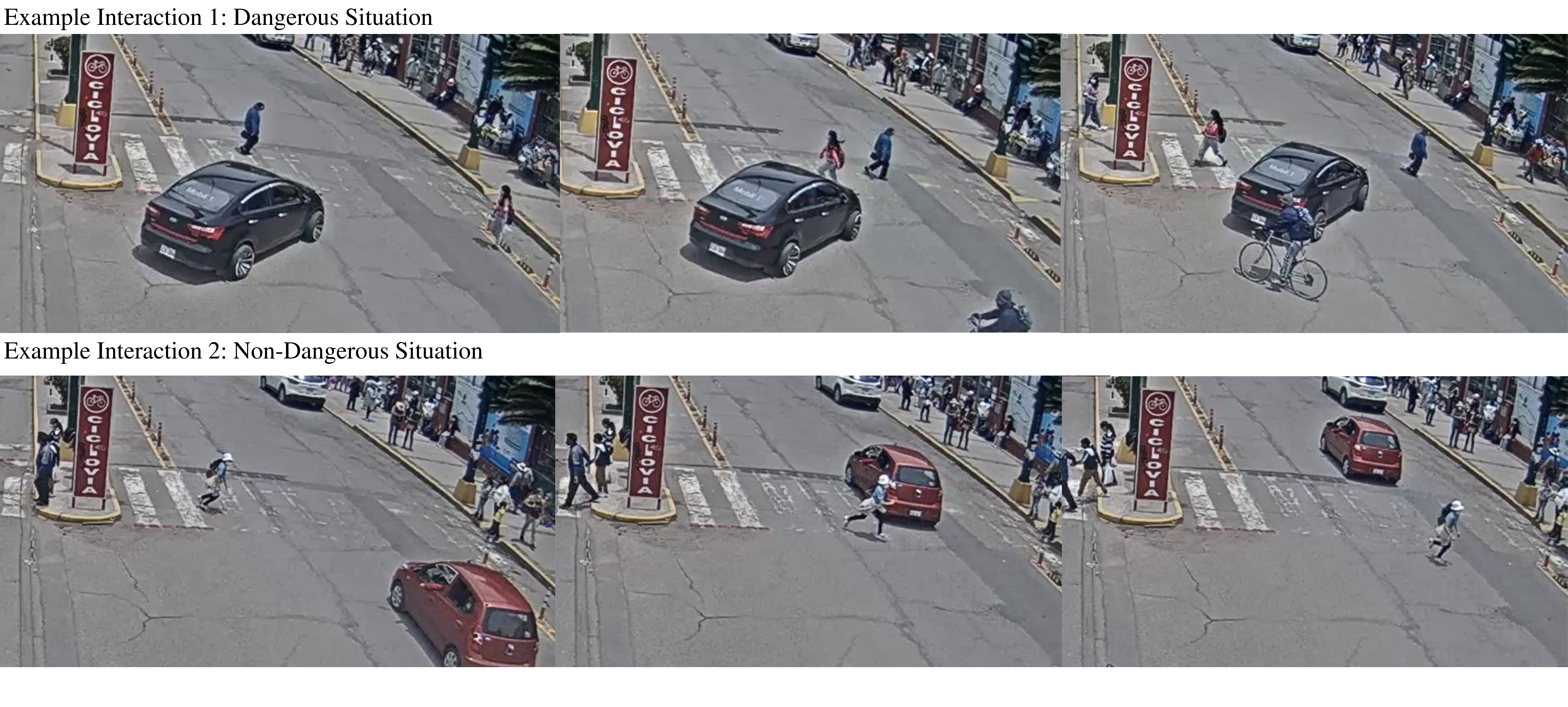}
\caption{\label{fig:danger}Vehicle-Pedestrian interactions captured by our dataset. Example interaction 1 shows a dangerous situation. Example Interaction 2 shows a non-dangerous situation where vehicle-pedestrian were very close.}
\end{figure}

\emph{I see you} captures vehicle-pedestrian avoidance behaviors in dangerous situations and scenarios where vehicle-pedestrian are very close but do not represent a dangerous situation (Figure \ref{fig:danger}) as this would provide valuable negative feedback. For each case, we provide processed vehicle and pedestrian trajectories in GPS coordinates.

\section{\emph{I see you} Dataset}
\label{sec:org72b2e83}

\begin{figure}[htbp]
\centering
\includegraphics[angle=0,width=12cm]{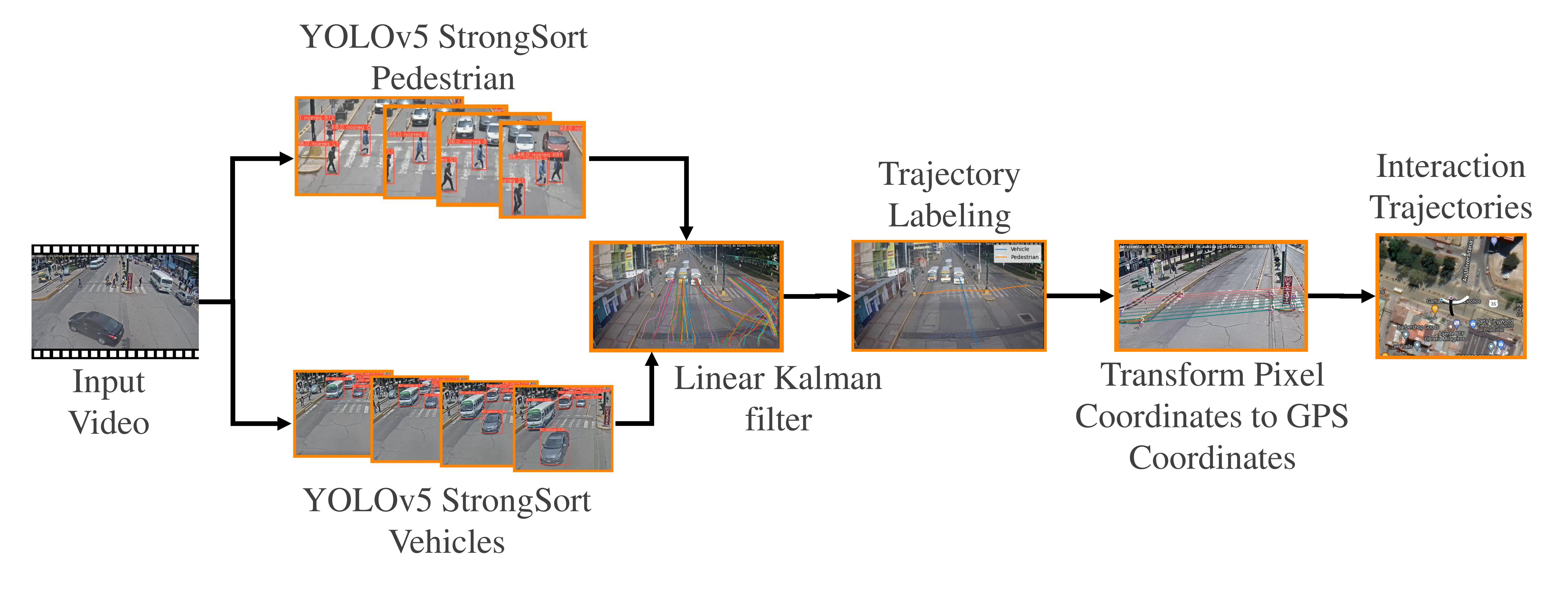}
\caption{\label{fig:flow}Pipeline for the proposed system.}
\end{figure}

We developed a pipeline to collect \emph{I see you} (Figure \ref{fig:flow}). The object detection task was performed using a YOLOv5 \cite{glenn_jocher_2020_4154370}, for tracking we used StrongSORT \cite{yolov5-strongsort-osnet-2022} then we used a linear Kalman filter \cite{yang_top-view_2019} to remove noise from the trajectories and finally for transformation to GPS coordinates we used Perspective-n-Point \cite{Tang17AIC}.

\begin{figure}[h]
        \begin{subfigure}[b]{0.5\textwidth}
            \centering
            \includegraphics[width=5.7cm]{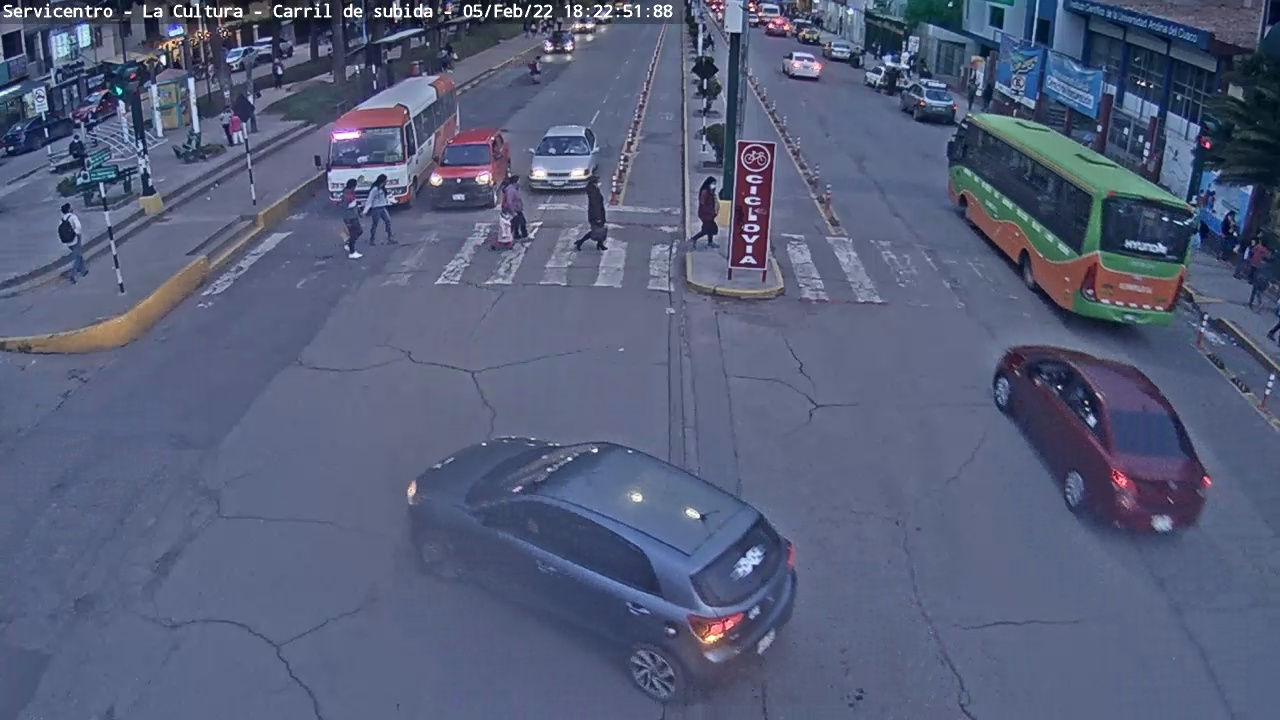}
            \caption{Servicentro}
            \label{fig:cam1}
        \end{subfigure}
        \begin{subfigure}[b]{0.5\textwidth}
            \centering
            \includegraphics[width=5.7cm]{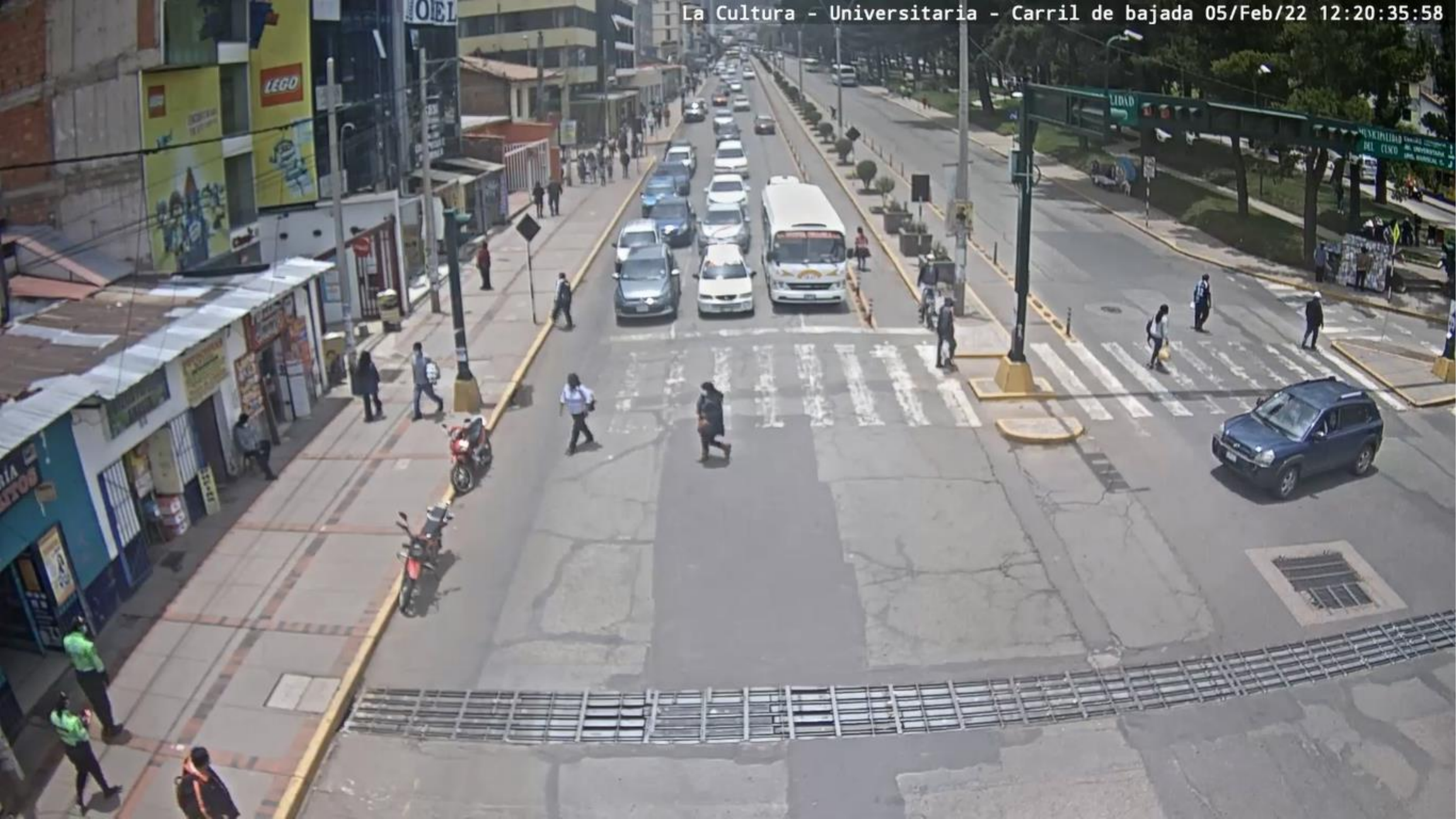}
            \caption{Univesitaria}
            \label{fig:cam2}
        \end{subfigure}
        \begin{subfigure}[b]{0.5\textwidth}
            \centering
            \includegraphics[width=5.7cm]{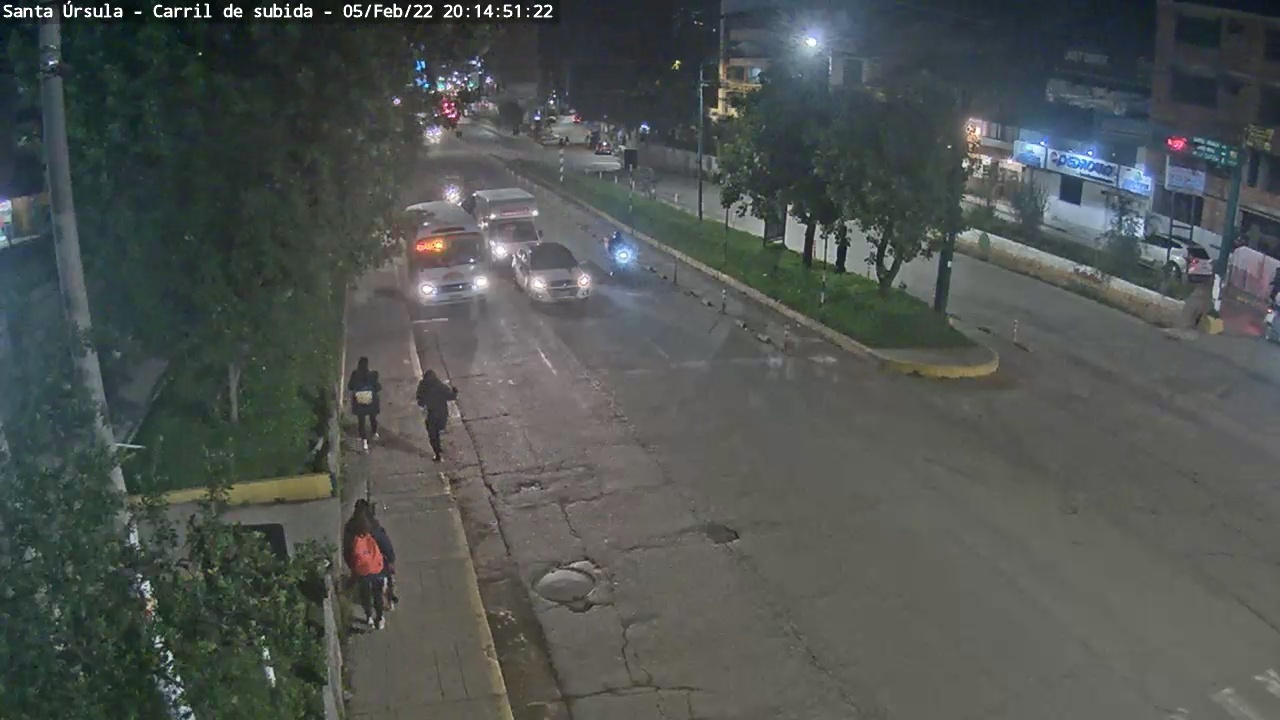}
            \caption{Santa Ursula}
            \label{fig:cam3}
        \end{subfigure}
        \begin{subfigure}[b]{0.5\textwidth}
            \centering
            \includegraphics[width=5.7cm]{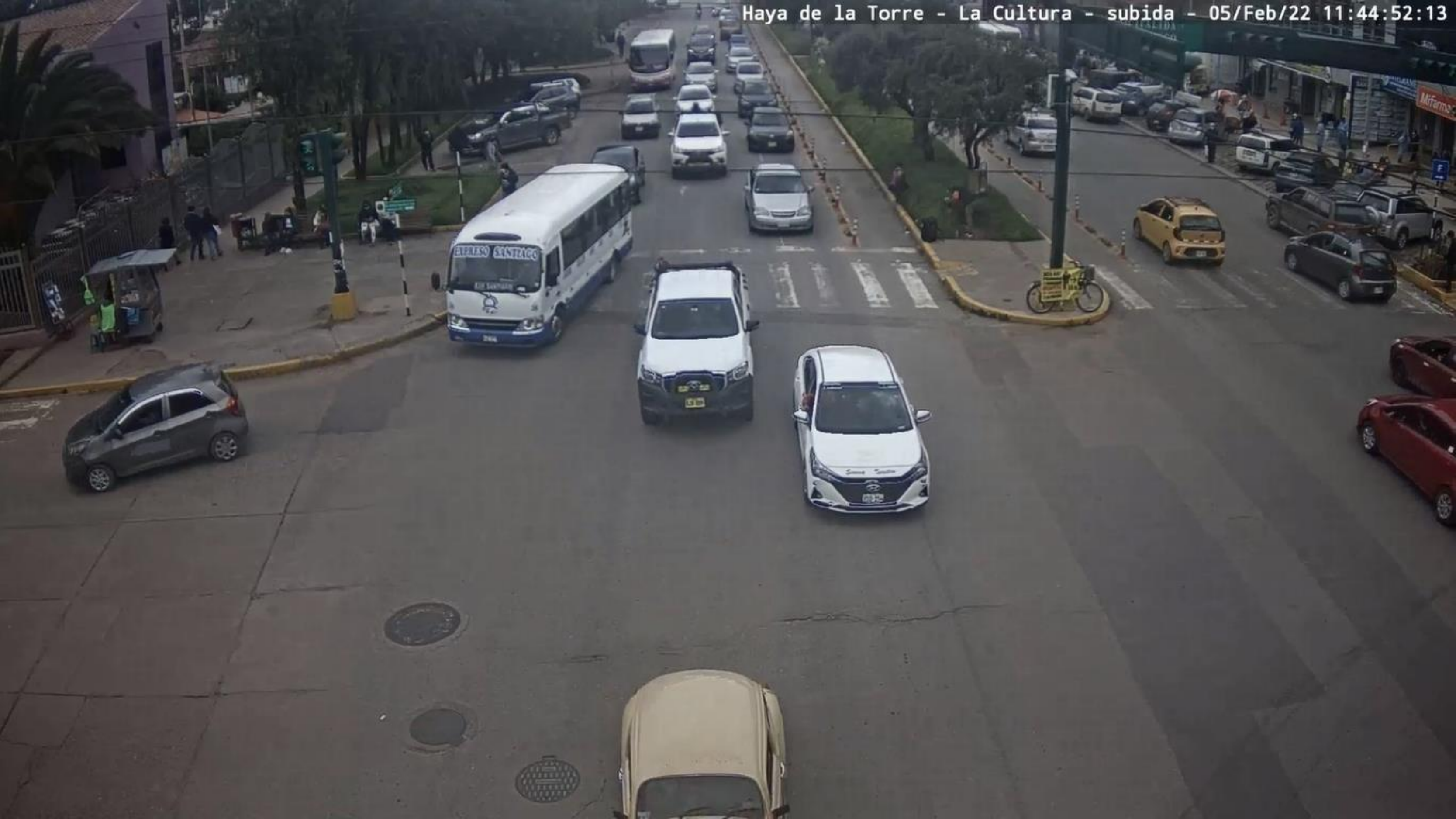}
            \caption{Haya de la Torre}
            \label{fig:cam4}
        \end{subfigure}
    \caption{Illumination variations of the collected footage in four intersections.}
    \label{fig:intersections}
\end{figure}

Video footage was collected from publicly available trafﬁc surveillance videos at seven signalized intersections in Cusco-Peru (Figure \ref{fig:intersections}). The seven intersections were selected near schools (Figure \ref{fig:cam1}), colleges (Figure \ref{fig:cam2}), churches (Figure \ref{fig:cam3}), and hospitals (Figure \ref{fig:cam4}) to ensure a diverse distribution in pedestrian characteristics (genre, age). These intersections also exhibit the transit from particular vehicles, taxis, and public transport buses where the latter show, particularly aggressive driving behaviors.

Each video is 18 hours long taken from 6 am to 11:59 pm on February 5, 2022. Every video has daytime, sunset and nighttime scenes these variations in illumination lead us to fine-tune the YOLOv5 models used for detection. For each category (vehicle, pedestrian) we used a different pre-trained YOLOv5 model, the vehicle category is detected by the model used in \cite{hu_turning_2021} and the pedestrian category is detected by a model pre-trained in the COCO dataset.

To capture all illumination variations we sampled the 18-hour videos saving a frame every 6 minutes resulting in 4246 images in total. Employing a \(90\%\) train, \(10\%\) test distribution, we obtained the results described in Table \ref{tab:yolo_results} for object detection in the test subset for each category.

\begin{table}[htbp]
\caption{\label{tab:yolo_results}YOLOv5 Training Results}
\centering
\begin{tabular}{lrrrr}
\hline
Category & P & R & mAP@0.5 & mAP@0.5:0.95\\
\hline
Vehicle & \(0.934\) & \(0.916\) & \(0.974\) & \(0.76\)\\
Pedestrian & \(0.864\) & \(0.853\) & \(0.91\) & \(0.595\)\\
\hline
\end{tabular}
\end{table}

\begin{figure}[htbp]
\centering
\includegraphics[angle=0,width=10cm]{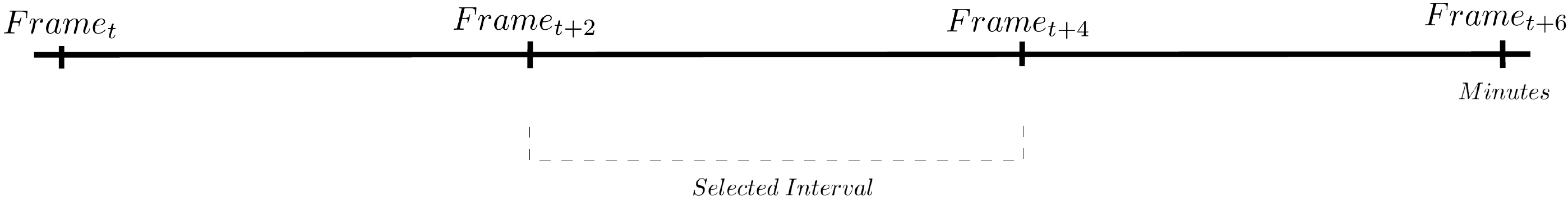}
\caption{\label{fig:sampling}Sampling approach used to save each 2-minute clip}
\end{figure}

Since the YOLOv5 models had already been trained on the scenes every 6 minutes we had to make sure that these same scenes were not presented to the detector on the tracking step, this required a new sampling approach. We noted that if a frame was taken when the traffic light was red, the scene remained almost the same until the traffic light turned green. In the worst-case scenario a frame could have been taken at the same time that the traffic light turned to red, this implies that a scene could remain almost the same for up to 90 seconds (maximum duration of red light). To overcome that problem we separated the 6 minutes interval between frames into 3 intervals 2 minutes long each, then we saved the intermediate 2-minute interval (Figure \ref{fig:sampling}) for the tracking step.

The tracking step was conducted using StrongSORT \cite{yolov5-strongsort-osnet-2022}, this tracker algorithm used the YOLOv5 models previously fine-tuned to detect the road users in the 2-minute clips and extract the trajectories of each category assigning them id numbers that were used in the manual labeling step. The resulting trajectories were noisy with spikes indicating unreal accelerations and speeds, this fact lead us to use a linear Kalman filter using the approach developed by \citet{yang_top-view_2019}.

The filtered trajectories were manually labeled with the id generated by the tracker of the vehicle and pedestrian involved in the near-accident scenario. The labeled trajectories were converted from pixel coordinates to GPS coordinates using semi-automatic camera calibration based on Perspective-n-Point (PnP) \cite{Tang17AIC} . Finally, we compare our dataset with other trajectory datasets used for pedestrian trajectory prediction in various cases (Table \ref{tab:dataset_compare}).

\begin{table}[htbp]
\caption{\label{tab:dataset_compare}Comparison with Other Public Available Trajectory Datasets}
\centering
\begin{tabular}{lllllll}
\hline
Dataset & Scenarios & Interaction & Agents & Method of & FPS & Amount\\
 &  & Type &  & Annotation &  & of Trajec-\\
 &  &  &  &  &  & tories\\
\hline
DUT \cite{yang_top-view_2019} & Campus & Vehicle-Crowd & Car, Pedestrian & CSRT tracker & 23.98 & 1793\\
 &  & in shared spaces &  &  &  & \\
\hline
\emph{I see you} & Signalized & Vehicle-Pedestrian & Car, Pedestrian & manual & 30 & 340\\
 & Intersections & in near-accident &  &  &  & \\
 &  & scenarios &  &  &  & \\
\hline
ETH \cite{pell2009} & Campus, & Pedestrians in & Pedestrian & manual & 2.5 & 650\\
 & Urban Street & busy scenarios &  &  &  & \\
\hline
UCY \cite{lerner_crowds_2007} & Urban Street, & multi-human & Pedestrian & manual & 2.5 & 909\\
 & Campus, Park & interaction scenarios &  &  &  & \\
\hline
\end{tabular}
\end{table}

\section{Statistics}
\label{sec:org6220385}

\begin{figure}[htbp]
\centering
\includegraphics[angle=0,width=10cm]{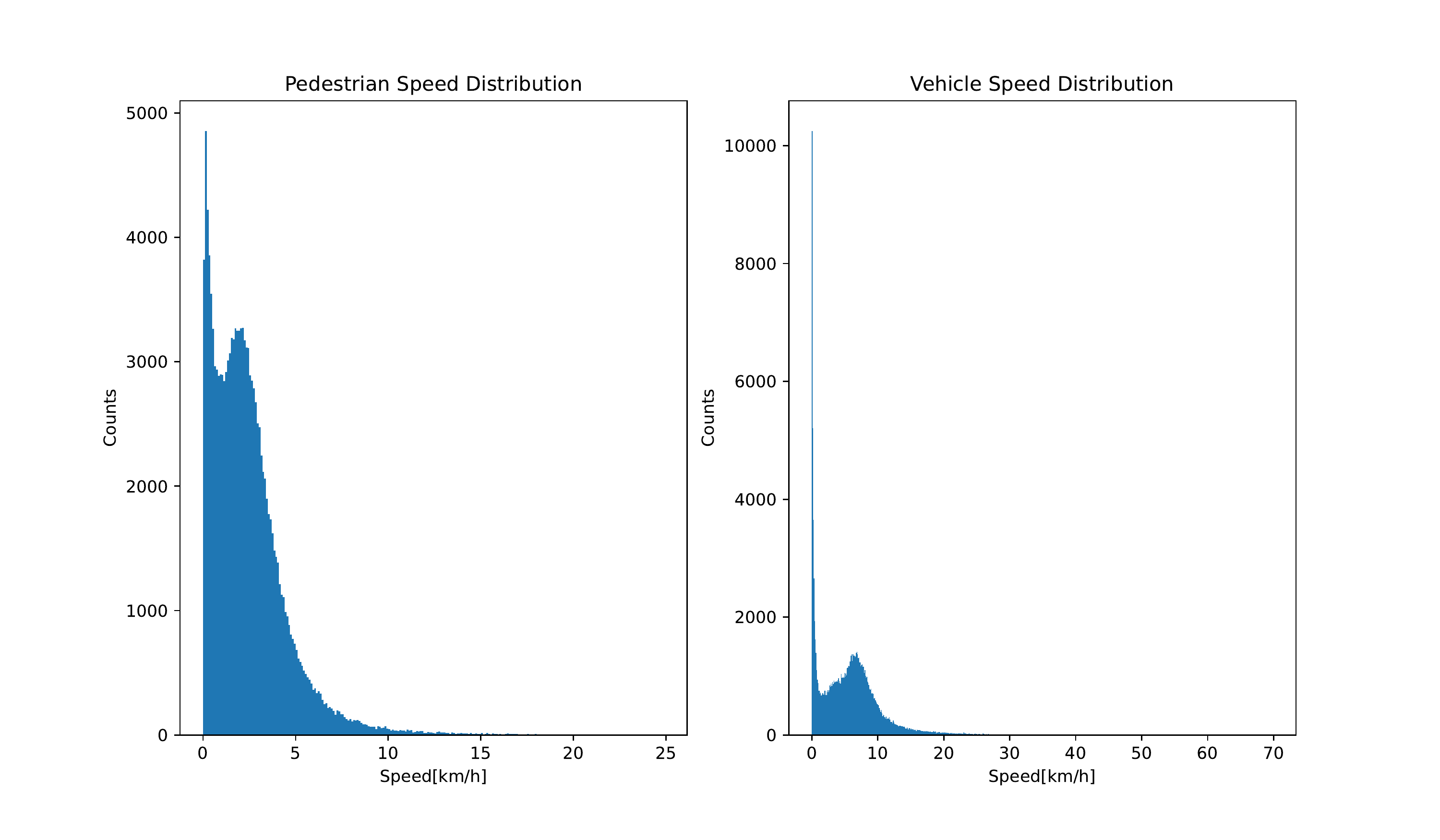}
\caption{\label{fig:data}Speed distributions for vehicle and pedestrian categories in \emph{I see you}}
\end{figure}

Speed is preferred to train pedestrian trajectory prediction models rather than absolute position \cite{becker_evaluation_2018,noh_novel_2022}. This is because it does not depend on the reference system and therefore it can be used as a more general way to describe motion. In Figure \ref{fig:data} speed distributions were calculated for vehicles and pedestrians. Both distributions show peaks at \(0\frac{km}{h}\) this is because vehicles and pedestrians are stopped by traffic lights at signalized intersections. Also, each distribution has a second peak that shows the mean speed of vehicles and pedestrians during the near-accident scenarios.

\section{Conclusions and Future Work}
\label{sec:org347ebe0}

In this work, we addressed the need for trajectory data in near-accident scenarios for which we have developed a pipeline for collecting the trajectories of pedestrians and vehicles. During this process, we have assessed the limitations of our pipeline in which manual labeling is a process that can be automated using clusters to select the trajectories that correspond to interactions in near accident scenarios and also could be expanded to other types of interactions by selecting the appropriate clusters for each type of interaction. Pedestrian risk-seeking behaviors could be similar in other cities, however, to use the developed pipeline in other locations it would be necessary to fine-tune the YOLOv5 models in those locations to get good detection results.
\section*{Acknowledgements}
\label{sec:org0615ea2}
We would like to thank the Data Analysis Laboratory(LAAD) for providing the GPU used for this project and Rodolfo Quispe for his continuous support and feedback.

\bibliographystyle{IEEEtranN}
\bibliography{I_see_you}
\end{document}